\documentclass{article}
\usepackage[OT1]{fontenc}
\usepackage{iftex}
\ifPDFTeX
\fi
\usepackage{iclr2027_conference,times}
\usepackage{amsmath}
\usepackage{amssymb}
\usepackage{booktabs}
\usepackage{graphicx}
\usepackage{xcolor}
\usepackage{url}
\usepackage{xspace}
\usepackage{float}
\usepackage{multirow}
\usepackage{needspace}
\usepackage{tikz}
\usetikzlibrary{arrows.meta}
\usepackage{hyperref}
\hypersetup{
  colorlinks=true,
  citecolor=teal,
  linkcolor=magenta,
  urlcolor=magenta
}
\newcommand{\sees}{\textsc{SEES}\xspace}

\title{SEES: A Self-Evolving Embodied System via Failure-Guided VLA Policy Adaptation}

\author{%
\begin{minipage}[t]{\dimexpr\textwidth-2\tabcolsep\relax}
\centering
Ziwen Li$^{1,*}$ \quad
Hanlue Zhang$^{1,*}$ \quad
Zhenyang Ren$^{2,*}$ \quad
Tianyu Huang$^{2,*}$ \\
Runqi Lin$^{3}$ \quad
Haoyu Wang$^{1}$ \quad
Zhengqing Gao$^{1}$ \quad
Yandong Guo$^{4}$ \\
Fakhri Karray$^{5,1}$ \quad
Tongliang Liu$^{2}$ \quad
Chris Russell$^{3}$ \quad
Mingming Gong$^{6,1}$ \\[3pt]
\normalfont\small
$^{1}$Mohamed bin Zayed University of Artificial Intelligence \\
$^{2}$University of Sydney \quad
$^{3}$University of Oxford \quad
$^{4}$AI$^2$ Robotics \\
$^{5}$University of Waterloo \quad
$^{6}$University of Melbourne \\[2pt]
$^{*}$These authors contributed equally and share first authorship.
\end{minipage}%
}

\hypersetup{
  pdftitle={SEES: A Self-Evolving Embodied System via Failure-Guided VLA Policy Adaptation},
  pdfauthor={Ziwen Li, Hanlue Zhang, Zhenyang Ren, Tianyu Huang, Runqi Lin, Haoyu Wang, Zhengqing Gao, Yandong Guo, Fakhri Karray, Tongliang Liu, Chris Russell, Mingming Gong}
}

\iclrfinalcopy

\begin{document}
\maketitle
\lhead{Preprint}
\vspace{-1.5em}
\begin{abstract}

 Recent vision-language-action (VLA) policies demonstrate promising generalization across diverse short-horizon tasks.
 However, they remain unreliable on long-horizon tasks, partly because the large-scale training data is biased toward single-stage manipulation tasks that are cheaper to demonstrate.
 A single weak atomic skill can cause failures across multiple multi-stage tasks.
 To address such failures, existing methods often require experts to identify the bottleneck and provide additional demonstrations, making the improvement costly and potentially impractical after deployment. To this end, we present a Self-Evolving Embodied System (SEES) that learns from failures and improves the VLA policy without additional expert demonstrations.
\sees decomposes long-horizon tasks into atomic tasks and routes them to corresponding family policies. Each family consists of related atomic skills that share one VLA adapter. During execution, the system automatically monitors atomic-task outcomes to identify the most frequently failing atomic skills as the current bottlenecks.
To overcome these bottlenecks, \sees constructs tailored RL tasks in simulation by restoring previously encountered states and generating task-specific success criteria with an LLM.
Online RL updates the shared family adapters to promote positive transfer among related atomic skills and cumulative improvement across evolution rounds.
Extensive experiments show that \sees can be integrated with different VLA backbones to progressively improve their long-horizon performance.
We also observe continued improvement on unseen tasks, providing evidence of transfer beyond the evolution settings.

\end{abstract}

\section{Introduction}

Recent VLA policies leverage large-scale vision-language pretraining and diverse robot trajectories to perform a broad range of robot tasks~\citep{zitkovich2023rt2,kim2024openvla,black2024pi0,physicalintelligence2025pi05,nvidia2025gr00tn1,openx2025rtx,walke2024bridgedatav2,khazatsky2025droid}. However, long-horizon execution remains unreliable, partly because collecting multi-stage demonstrations is costly and many existing robot datasets focus on short-horizon manipulation~\citep{garrett2024skillmimicgen,ceola2024lhmanip}. A household mobile-manipulation task may require multiple navigation and object-interaction stages, and full-task success depends on completing every required stage~\citep{ahn2022saycan,gu2022m3,liang2022codeaspolicies}. Consequently, weaknesses in a few shared atomic capabilities can cause failures across a wide range of long-horizon tasks.

One way to address these failures is through human-in-the-loop policy improvement. Experts review performance across tasks and inspect failed executions to identify bottlenecks, then collect targeted demonstrations to improve weak capabilities and overall task success. Intervention-based methods aim to use human effort more efficiently by having operators take over only when the policy is at risk of failure~\citep{kelly2019hgdagger,mandlekar2020human}. To further reduce the need for constant human monitoring, some methods automatically detect potential problems and request human intervention~\citep{hoque2021thriftydagger,liu2024runtime,liu2024siriusfleet}. These advances make human supervision more efficient, but collecting sufficient data for supervised fine-tuning (SFT) still requires substantial human effort.
Moreover, for commercial robots deployed in users' homes, repeatedly collecting data for specific tasks and environments may be impractical, limiting continued improvement when expert support is unavailable.
We therefore ask whether a robot can autonomously identify its bottlenecks and improve its policy accordingly without additional expert demonstrations.

\begingroup
\setlength{\intextsep}{\dimexpr\baselineskip-\parskip\relax}
\begin{figure}[H]
  \centering
  \includegraphics[width=0.95\textwidth,trim=0 10bp 0 0,clip]{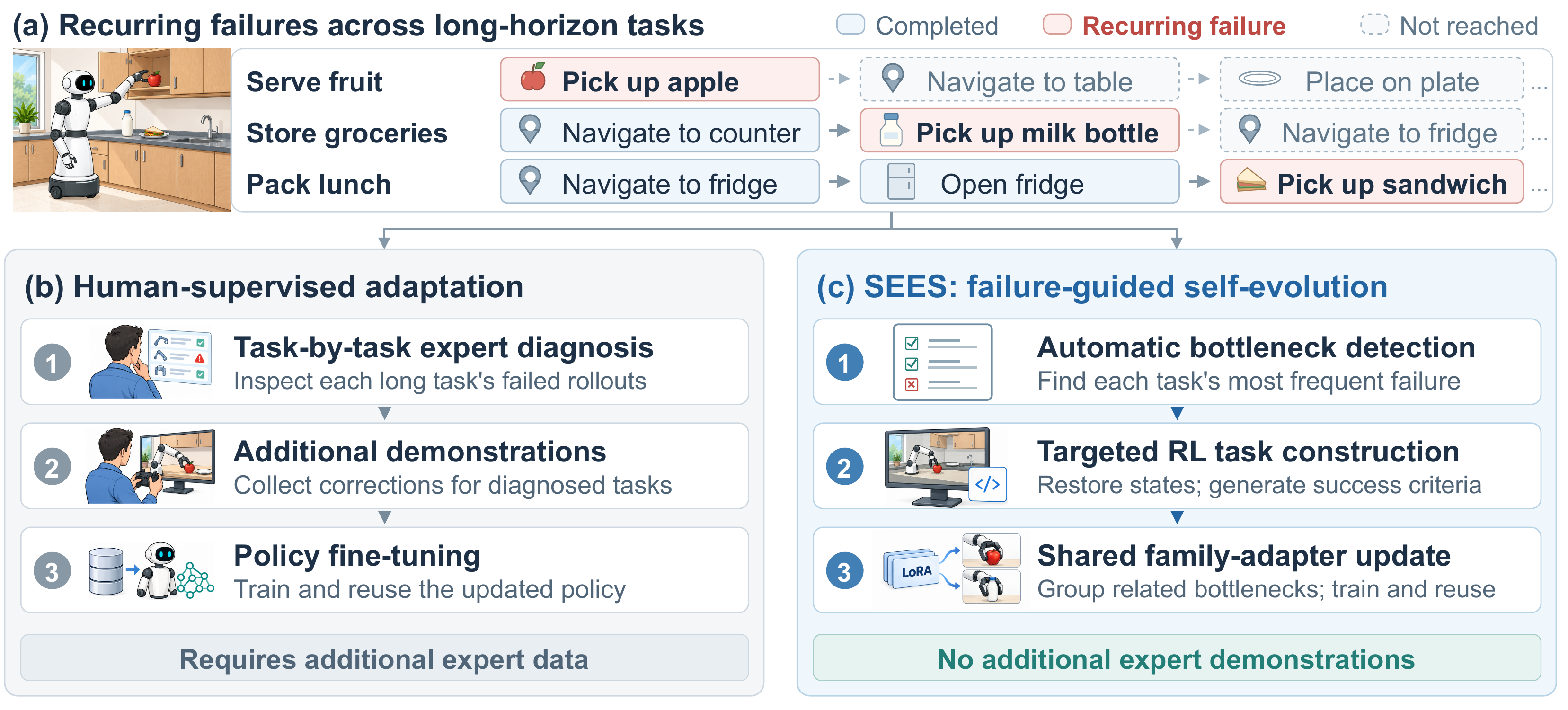}
\caption{\textbf{Human-supervised adaptation versus \sees self-evolution.}
(a) Related atomic failures can block progress across different long-horizon tasks.
(b) Human-supervised adaptation relies on expert diagnosis and additional demonstrations.
(c) \sees automatically identifies recurring bottlenecks, constructs targeted RL tasks by restoring encountered states and generating success criteria, and updates shared family adapters.}
  \label{fig:motivation}
\end{figure}
\vspace{-2pt}
\endgroup

We introduce \sees, a self-evolution framework that can be integrated with different VLA backbones (Fig.~\ref{fig:motivation}).
It decomposes long-horizon tasks into atomic tasks and executes each using the VLA adapter shared by its skill family. During execution, a learned termination head provides completion signals for timely stopping and outcome monitoring. By aggregating atomic-task outcomes across executions, \sees identifies the bottlenecks that most frequently prevent full-task completion.

For each bottleneck, \sees uses an LLM-based task builder to automatically construct a targeted RL task in simulation. The builder restores previously encountered states as initial conditions and generates task-specific success criteria for computing sparse rewards.
Rather than training a separate adapter for each atomic task, \sees jointly trains a shared adapter on bottlenecks from the same family, allowing related tasks to benefit even when they are not selected for RL.
We use low-rank adaptation (LoRA) to store these family-level updates without maintaining a full VLA model for each family~\citep{hu2022lora}. Checkpoints are selected for reuse based on their validation gains on the chosen bottleneck tasks. Repeating this evolution cycle progressively strengthens weak atomic skills and improves overall performance on long-horizon tasks.

We evaluate \sees on RoboCasa365~\citep{nasiriany2026robocasa365} using GR00T N1.5~\citep{nvidia2025gr00tn15} and RLDX-1~\citep{kim2026rldx1}, and further examine it across all four LIBERO suites~\citep{liu2023libero}. On RoboCasa365, both backbones achieve higher atomic and long-horizon success after three evolution rounds. More notably, on Composite-Unseen, which contains entirely new tasks, success increases from 3.5\% to 6.1\% for GR00T N1.5 and from 6.0\% to 7.1\% for RLDX-1. On LIBERO, failure-guided policy evolution improves success across all four suites, raising the average from 53.1\% to 65.4\% after three rounds and benefiting tasks never selected for evolution. Further analyses validate termination-based failure attribution, failure-based bottleneck selection, and cross-task transfer through family-level adaptation. Together, these results demonstrate the effectiveness of \sees across different VLA backbones and benchmarks, as well as its ability to generalize to unseen tasks.
This paper makes three main contributions:
\begingroup
\setlength{\topsep}{3pt}
\setlength{\itemsep}{4pt}
\setlength{\parsep}{0pt}
\begin{itemize}
    \item We introduce \sees, a self-evolving embodied system compatible with different VLA backbones. It automatically identifies atomic bottlenecks and constructs targeted simulation RL tasks for policy improvement without additional expert demonstrations.
    \item We develop a failure-guided family-level evolution mechanism that jointly adapts related bottlenecks selected from long-horizon execution failures. The resulting updates strengthen capabilities shared across the family and can benefit its atomic skills collectively.
    \item Experiments across VLA backbones and benchmarks demonstrate progressive policy improvement, with gains extending to unseen tasks. Targeted analyses further validate the effectiveness of failure attribution, bottleneck selection, and family-level adaptation.
\end{itemize}
\endgroup

\section{Related Work}

\noindent\textbf{Autonomous Robot Self-Improvement.}
A shared goal is to let robots improve from execution experience without repeatedly collecting expert demonstrations.
ASPIRE repairs robot programs and accumulates reusable skills, but the behaviors it can express remain constrained by the available perception and control primitives~\citep{lu2026aspire}.
Q-Planning learns from deployment rollouts to improve action selection, but relies on its frozen policy to propose useful actions~\citep{giridhar2026qplanning}.
Direct VLA adaptation offers a complementary route by using execution feedback to improve action generation, rather than only program structure or action selection.

Other systems improve policies through autonomous data collection.
SOAR selects feasible tasks to diversify experience, while RoboClaw repeatedly collects trajectories to refine manipulation policies~\citep{zhou2025soar,li2026roboclaw}.
These approaches reduce the effort needed to acquire training data.
Broader experience, however, does not necessarily prioritize the skills that most frequently prevent long-horizon tasks from succeeding.
This distinction matters when limited training resources must be allocated across many skills.

The cost of repeated practice is another consideration.
RISE and VLAW reduce physical interaction through world-model rollouts, which require learned models that reliably predict interactions and their outcomes~\citep{yang2026rise,guo2026vlaw}.
Given an existing simulator, \sees records states during execution and restores them later for targeted training based on accumulated failure statistics.
The resulting updates improve subsequent executions.

\noindent\textbf{Targeted RL Task Construction.}
RLinf-VLA and SimpleVLA-RL fine-tune VLA policies with online RL on given tasks and reward definitions~\citep{zang2026rlinfvla,li2026simplevlarl}.
Eureka and Text2Reward generate reward programs, RoboGen and OMNI-EPIC generate tasks and environments, and ReinforceGen combines long-horizon task decomposition, automated data generation, and RL~\citep{ma2024eureka,xie2024text2reward,wang2024robogen,faldor2024omni,zhou2026reinforcegen}.
SAFE detects failures, SeqVLA predicts completion, and BATON localizes stage failures without updating the policy~\citep{gu2025safe,yang2025seqvla,xu2026baton}.
\sees aggregates stage-level outcomes across rollouts to choose which atomic skill to train, restores states encountered before that skill, and generates a Boolean success predicate to define its sparse reward.

\noindent\textbf{Skill-Structured Policy Adaptation.}
Modular adaptation balances preserving existing capabilities against sharing improvements.
CORAL isolates task-specific LoRA experts to prevent parameter interference, but an update to one expert does not directly improve the others~\citep{luo2026coral}.
CLARE reuses and expands adapters, while Stellar VLA organizes shared knowledge through task--skill structure~\citep{romer2026clare,wu2025stellarvla}.
OrthoSkillVLA constrains parameter updates and maintains skill-specific experts to protect previously learned skills~\citep{wang2026orthoskillvla}.
These methods primarily address how to incorporate new task or skill data while retaining previous abilities.
\sees builds on structured parameter sharing but determines its training targets from recurring execution failures rather than incoming demonstration datasets.
Related bottlenecks jointly train one family adapter, and the updated adapter is reused by all tasks assigned to that family.
This allows targeted practice to benefit related tasks not selected for RL, without maintaining and training a separate adapter for every atomic task.

\section{Method}
\label{sec:method}

\begin{figure}[!tbp]
  \centering
  \includegraphics[width=0.95\textwidth]{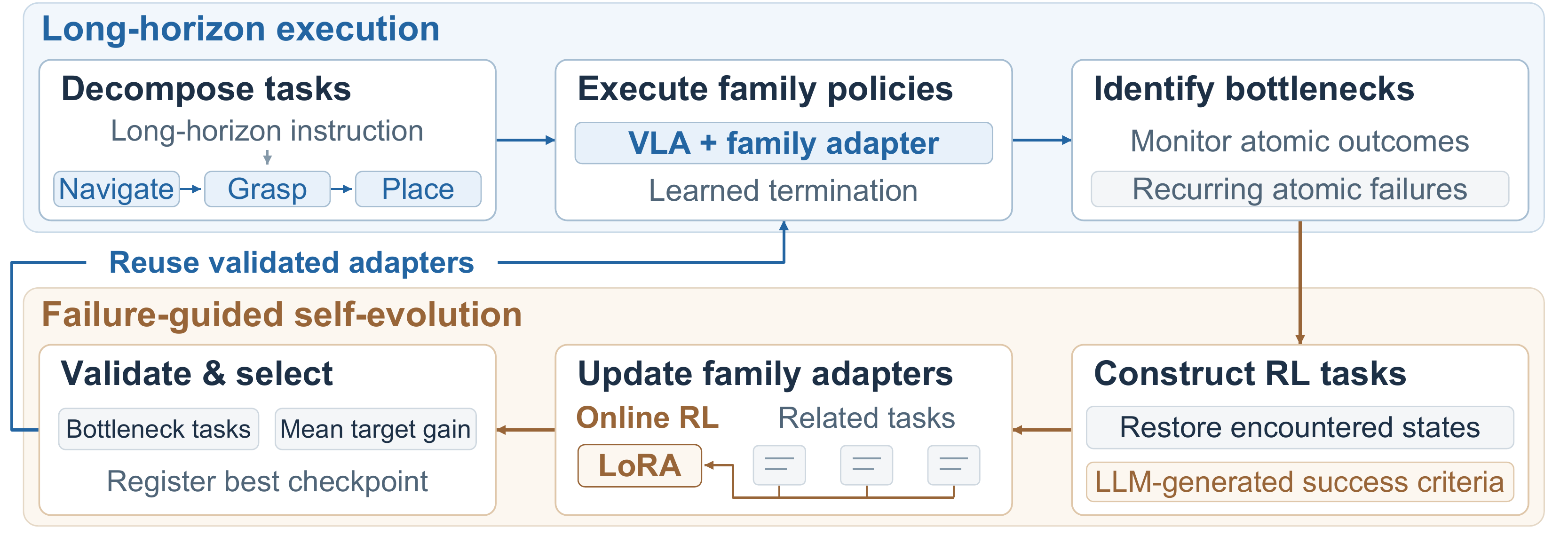}
  \caption{\textbf{The \sees execution--evolution loop.}
  In the top row, \sees decomposes long-horizon instructions, executes atomic tasks with family policies, and identifies bottlenecks using learned termination and outcome statistics.
  In the bottom row (right to left), restored states and LLM-generated success criteria define targeted simulation RL tasks.
  Online RL jointly trains each family adapter on its selected bottleneck tasks. The checkpoint with the largest positive gain on bottleneck-task validation is registered and returned to execution.
}
  \label{fig:overview}
  \vspace{-12pt}
\end{figure}

\sees follows a repeated execution-and-adaptation loop (Fig.~\ref{fig:overview}). It first decomposes long-horizon tasks, executes them with family policies, and aggregates atomic outcomes to identify recurring bottlenecks (Sec.~\ref{sec:termination}). For each selected bottleneck, an LLM-based task builder restores encountered states and generates a success function to construct a targeted simulation RL task (Sec.~\ref{sec:rl-task-construction}). Online RL then updates the shared adapters of the affected skill families, and the checkpoint with the largest positive target-task gain is registered for subsequent execution (Sec.~\ref{sec:family-evolution}).

\subsection{Execution and Bottleneck Identification}
\label{sec:termination}

Given a long-horizon instruction $c$ and initial-scene observations, a frozen VLM planner produces an ordered sequence of executable atomic tasks $\mathcal{A}_c$. A language agent maps each atomic task $i$ to a predefined skill family $g(i)$ according to the capability it requires. Tasks within a family retain their individual instructions but share the same VLA adapter, allowing related tasks from different composite plans to reuse the family's current policy. We denote the frozen pretrained VLA parameters by $\theta$ and the adapter shared by family $f$ by $\phi_f$. The policy $\pi_{\theta,\phi_{g(i)}}$ executes task $i$ conditioned on its atomic instruction and current observations. Before an updated adapter becomes available for a family, its tasks are executed by the pretrained base VLA. Appendix~\ref{app:vlm-decomposition} provides the planner configuration and skill-family definitions.

To stop completed subtasks promptly and provide stage boundaries for failure attribution, we attach a termination head to the pretrained VLA and train it to predict atomic task completion, while keeping all original parameters frozen. During long-horizon task execution, a predicted termination ends the current motion and allows the next atomic task to begin. A subtask is marked as failed if no termination signal is emitted before the step limit or unsuccessful environment termination. The audit in Appendix~\ref{app:termination-audit} supports using termination signals for failure attribution. We then aggregate these outcomes across rollouts to identify recurring bottlenecks.

For each failed rollout of composite task $c$, we assign the failure to the first reached atomic stage $i$ that prevents execution from continuing. Downstream stages blocked by this failure are not counted. Let $F_{c,i}$ denote the number of failed rollouts assigned to stage $i$. At each evolution round, we select the stage with the highest failure count as the bottleneck $b_c$ for each composite task:
\begin{equation}
b_c=\arg\max_{i\in\mathcal{A}_c}F_{c,i}.
\label{eq:bottleneck-selection}
\end{equation}
We select one bottleneck per composite task each round to focus the fixed adaptation budget on its most frequent blocker. Selecting bottlenecks before grouping them by skill family keeps the decision sensitive to specific weak stages, even when other tasks in the family are reliable.

\subsection{Bottleneck-Targeted RL Task Construction}
\label{sec:rl-task-construction}

For each selected bottleneck task $i$, the RL task builder supplies two components: initial states that reproduce its execution context and a success predicate for sparse rewards. We collect candidate states immediately before $i$ begins in composite-task rollouts, whether the attempt subsequently succeeds or fails. Restoring the scene configuration and task-dependent episode state allows training to start directly at the bottleneck without replaying earlier stages (Appendix~\ref{app:state-restoration}).

An LLM generates the success predicate $C_i(s)$, a Boolean success-checking function, from the task instruction $l_i$, the identifiers of relevant objects and fixtures, and documentation for simulator functions that query their states and relations. The function returns whether the atomic task is complete and is used to compute sparse rewards.  States that cannot be restored or already satisfy $C_i$ are excluded, leaving the initialization set $\mathcal{I}_i$. Appendix~\ref{app:criterion-generation} provides further details and separately evaluates agreement with official success predicates.
Reward construction therefore relies on the generated function and documented simulator interfaces rather than privileged task oracles, which supports extending the same pipeline to simulators reconstructed from real-world execution traces.

We keep the simulator dynamics unchanged and define the RL task for atomic task $i$ as
\begin{equation}
\mathcal{T}_i = (l_i, \mu_i^0, r_i),
\qquad
\mu_i^0 = \operatorname{Uniform}(\mathcal{I}_i),
\label{eq:rl_task}
\end{equation}
where $\mu_i^0$ samples an episode's initial state from the restored states. Let $S_{i,t}$ indicate whether task success has been confirmed by time $t$ according to $C_i$. Once confirmed, $S_{i,t}$ remains one for the rest of the episode. With $S_{i,0}=0$, the sparse reward is
\begin{equation}
r_{i,t} = S_{i,t} - S_{i,t-1}.
\label{eq:rl_reward}
\end{equation}
Thus, reward is issued only upon the first confirmed completion. During RL training, $C_i$ determines task success and episode termination. During composite-task execution, the learned termination head instead determines when to hand off to the next atomic task. These constructed tasks provide the targets for the family-level adapter updates described in Sec.~\ref{sec:family-evolution}.

\subsection{Family-Level Policy Evolution}
\label{sec:family-evolution}

For each family containing one or more selected bottlenecks in round $k$, \sees forms a training pool $\mathcal{T}_{f,k}$ from the current target tasks and tasks selected in earlier rounds, removing duplicates. It samples tasks and their restored initial states from this pool to collect fresh rollouts for updating a single adapter shared by the family. Previously selected tasks therefore contribute new RL experience rather than stored trajectories. The sampling weights account for the number of valid initial states and gradually reduce the weight of older tasks (Appendix~\ref{app:family-sampling}). Families with no selected bottleneck in the current round are not updated.

The family adapter accumulates updates from related bottlenecks across evolution rounds. Since all tasks in the family use these parameters, improvements learned from the selected bottlenecks can also benefit related tasks that receive no direct RL training.

For each adapted action-head layer, we parameterize the family-specific weights as
\begin{equation}
W_f = W_0 + \frac{\alpha}{r} B_f A_f,
\label{eq:family-lora}
\end{equation}
where $W_0$ is frozen, $A_f$ and $B_f$ are trainable rank-$r$ factors, and $\alpha$ is the scaling coefficient~\citep{hu2022lora}. The LoRA factors, together with the family-specific value and termination heads, form the trainable parameters $\phi_f$ saved with the family checkpoint. Each round starts from the latest accepted checkpoint for that family, or from the prepared base VLA for its first update.

We optimize the family policy on the constructed RL tasks using PPO~\citep{schulman2017ppo}, with the training pipeline implemented on top of RLinf~\citep{yu2025rlinf}. Let $\ell_t(\phi_f)$ be the log-probability of a sampled policy transition, and let $\phi_f^{\mathrm{old}}$ denote the parameters that generated the rollout batch. The probability ratio and clipped PPO objective are
\begin{align}
\rho_t(\phi_f)
&=
\exp\!\left[
\ell_t(\phi_f)-\ell_t(\phi_f^{\mathrm{old}})
\right],
\label{eq:family-ratio}\\
J_f^{\mathrm{PPO}}(\phi_f)
&=
\mathbb{E}_t
\left[
\min\!\left(
\rho_t(\phi_f)\hat A_t,\,
\operatorname{clip}\!\left(
\rho_t(\phi_f),1-\epsilon,1+\epsilon
\right)\hat A_t
\right)
\right],
\label{eq:family-ppo}
\end{align}
where $\hat A_t$ is the estimated advantage and $\epsilon$ is the clipping threshold. Value regression and auxiliary termination supervision train the corresponding heads, while the pretrained VLA backbone and other families' adapters remain frozen. For flow-based policies, the likelihood refers to sampled stochastic denoising transitions rather than the marginal likelihood of the final action~\citep{zhang2025reinflow,chen2025pirl}. Optimization details are provided in Appendix~\ref{app:policy-optimization}.

After each family update, \sees evaluates the policy before the update and the saved checkpoints on the selected bottleneck tasks using the protocol in Appendix~\ref{app:checkpoint-validation}. Let $\mathcal{C}$ denote the set of saved checkpoints, and let $\Delta_T^{(j)}$ denote the change in the equally task-weighted mean success rate for checkpoint $j$ relative to the pre-update policy. We select the checkpoint with the largest gain:
\begin{equation}
j^* = \arg\max_{j\in\mathcal{C}} \Delta_T^{(j)},
\label{eq:target-checkpoint-selection}
\end{equation}
and register it if $\Delta_T^{(j^*)}>0$. Otherwise, we retain the current family policy (a positive-gain checkpoint was found in every RoboCasa365 evolution round).

Accepted family checkpoints are used in subsequent executions, whose outcomes provide evidence for the next round of bottleneck selection. Full composite-task evaluation measures the system-level effect of adaptation but is not an additional checkpoint-acceptance criterion. Composite-Unseen tasks remain excluded from adaptation and checkpoint selection.

\section{Experiments}
\label{sec:experiments}

\begin{figure}[!tbp]
  \centering
  \includegraphics[width=1.0\textwidth]{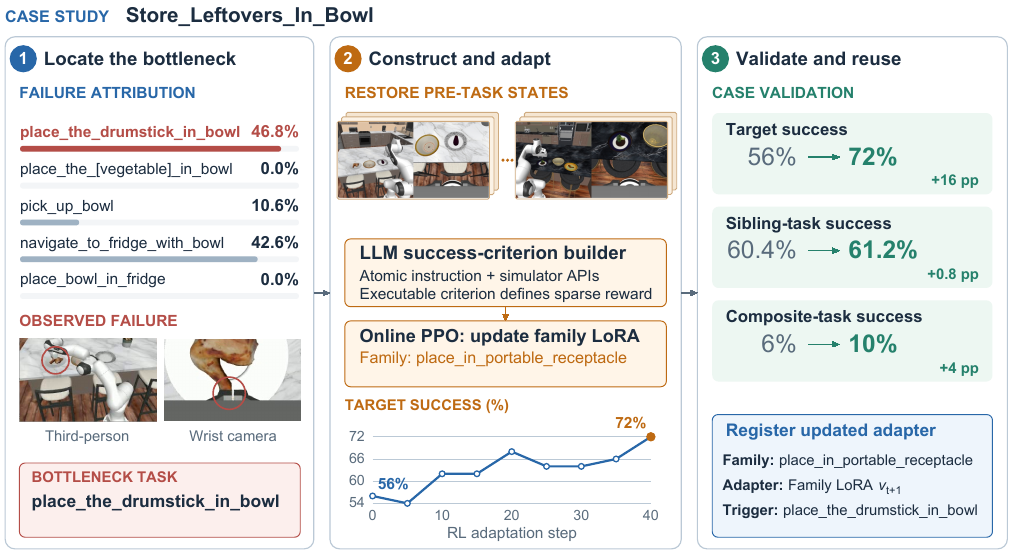}
\caption{\textbf{Representative \sees evolution case.}
\sees identifies an atomic bottleneck from composite-task failures,
constructs a targeted RL task from restored pre-task states, and updates
the corresponding family adapter. The resulting update improves the
bottleneck task, with gains extending to sibling task and composite task
execution.}
  \label{fig:evolution-transfer}
  \vspace{-10pt}
\end{figure}

\vspace{-5pt}
We conduct extensive experiments to evaluate the effectiveness and generalization of \sees and examine its key design choices. After introducing the experimental setup (Sec.~\ref{sec:experimental-setup}), we assess progressive policy improvement and generalization on RoboCasa365 with two VLA backbones (Sec.~\ref{sec:robocasa-evolution}) and further evaluate policy evolution across four LIBERO suites (Sec.~\ref{sec:libero-object}). We then compare family-level and atomic-specific adaptation to examine cross-task transfer (Sec.~\ref{sec:family-transfer}), and test whether prioritizing bottlenecks improves adaptation efficiency under a matched training budget (Sec.~\ref{sec:bottleneck-ablation}). Appendix~\ref{app:termination-audit} audits termination-based failure attribution, while Appendix~\ref{app:criterion-generation} evaluates the generated success functions against official predicates.

\vspace{-10pt}
\subsection{Experimental Setup}
\label{sec:experimental-setup}

\vspace{-5pt}

\noindent\textbf{Benchmarks.}
Our primary benchmark is RoboCasa365~\citep{nasiriany2026robocasa365}, which comprises 65 atomic and 300 composite kitchen manipulation tasks. We follow its 50-task evaluation protocol, covering Atomic (18 tasks), Composite-Seen (16 tasks), and Composite-Unseen (16 tasks). Atomic and Composite-Seen tasks are included in the RoboCasa365 pretraining data, while Composite-Unseen tasks are held out. We additionally evaluate policy evolution across the Object, Spatial, Goal, and Long suites of LIBERO~\citep{liu2023libero}.

\noindent\textbf{Backbones and initialization.}
On RoboCasa365, we evaluate GR00T N1.5~\citep{nvidia2025gr00tn15}, which combines a vision-language encoder with a diffusion-transformer action model, and RLDX-1~\citep{kim2026rldx1}, which uses a Multi-Stream Action Transformer. Both start from checkpoints trained on Human300 through supervised fine-tuning, using 100 demonstrations per task across 65 atomic and 235 composite tasks. For LIBERO, each suite starts from its corresponding GR00T N1.5 SFT checkpoint released by RLinf~\citep{yu2025rlinf}.

\noindent\textbf{Baselines and metrics.}
On RoboCasa365, we compare each reproduced base policy with execution-only \sees and \sees after one, two, and three evolution rounds. The execution-only configuration uses task decomposition, skill-family routing, and learned termination without RL adaptation. Comparison with the base policy measures the overall benefit of \sees, while comparison with execution only isolates the gains from policy evolution. Tab.~\ref{tab:main-results} also includes reference results from the original RoboCasa365 paper, marked with an asterisk. On LIBERO, we compare the initial SFT policies with those retained after three evolution rounds. We report success rates (\%), measuring standalone task completion for atomic tasks and full-task completion for composite tasks.

\begin{table}[t]
  \caption{
    \textbf{RoboCasa365 results.}
    Success rates (\%) across two VLA backbones.
    Starred results are taken from the original RoboCasa365
    paper. We evaluate all other results ourselves.
    Bold indicates the best result within each backbone.
  }
  \label{tab:backbone-validation}
  \label{tab:main-results}
  \vspace{3pt}
  \centering
  \small
  \setlength{\tabcolsep}{5pt}
  \renewcommand{\arraystretch}{1.05}

  \newcommand{\seesevolutionbranch}{%
    \makebox[2.25em][l]{%
      \hspace*{1.45em}%
      \smash{\tikz[baseline=-0.45ex]{%
        \draw[-{Stealth[length=0.9mm,width=0.65mm]},line width=0.3pt]
          (0,1ex) |- (0.7em,0);
      }}%
    }%
  }

  \begin{tabular*}{\linewidth}{
    @{\extracolsep{\fill}}llccc@{}
  }
    \toprule
    \multirow{2}{*}{Model}
      & \multirow{2}{*}{Configuration}
      & \multirow{2}{*}{Atomic}
      & \multicolumn{2}{c}{Composite} \\
    \cmidrule(lr){4-5}
      & & & Seen & Unseen \\
    \midrule

    Diffusion Policy$^{*}$
      & Base policy
      & 15.7 & \,\,0.2 & \,\,1.3 \\
    $\pi_0^{*}$
      & Base policy
      & 36.3 & \,\,5.2 & \,\,0.7 \\
    $\pi_{0.5}^{*}$
      & Base policy
      & 39.6 & \,\,7.1 & \,\,1.2 \\

    \midrule

    \multirow{5}{*}{GR00T N1.5}
      & Base policy
      & 47.2 & 10.2 & \,\,3.5 \\
      & \sees (execution only)
      & 47.2 & \,\,9.5 & \,\,4.7 \\
      & \seesevolutionbranch{\footnotesize +}\,\textit{1 evolution round}
      & \textbf{56.2} & 10.4 & \,\,5.2 \\
      & \seesevolutionbranch{\footnotesize +}\,\textit{2 evolution rounds}
      & 54.5 & 11.3 & \,\,5.7 \\
      & \seesevolutionbranch{\footnotesize +}\,\textit{3 evolution rounds}
      & 55.7 & \textbf{13.1} & \,\,\textbf{6.1} \\

    \midrule

    \multirow{5}{*}{RLDX-1}
      & Base policy
      & 66.4 & 20.4 & \,\,6.0 \\
      & \sees (execution only)
      & 66.4 & 17.6 & \,\,4.2 \\
      & \seesevolutionbranch{\footnotesize +}\,\textit{1 evolution round}
      & 68.5 & 19.9 & \,\,5.5 \\
      & \seesevolutionbranch{\footnotesize +}\,\textit{2 evolution rounds}
      & 68.9 & 20.4 & \,\,6.3 \\
      & \seesevolutionbranch{\footnotesize +}\,\textit{3 evolution rounds}
      & \textbf{70.3} & \textbf{21.8} & \,\,\textbf{7.1} \\

    \bottomrule
  \end{tabular*}
  \vspace{-10pt}
\end{table}

\subsection{Self-Evolution on RoboCasa365}
\label{sec:robocasa-evolution}

For each evolution round, we collect 50 rollouts for each of the 16 Composite-Seen tasks and allocate 40 PPO training steps to each active family. We evaluate the updated system using the same evaluation settings as the original RoboCasa365 paper~\citep{nasiriany2026robocasa365}. Composite-Unseen tasks remain entirely excluded from evolution and checkpoint selection. Optimization and checkpoint-validation details are provided in Appendices~\ref{app:policy-optimization} and~\ref{app:checkpoint-validation}.

Tab.~\ref{tab:main-results} shows that composite success improves at every evolution round for both backbones. Before evolution, however, execution-only \sees performs below the base policies on Composite-Seen. Our manual inspection finds that many decomposed subtasks have no standalone counterpart in the atomic portion of the RoboCasa365 pretraining data. Although the full composite tasks are seen during training, decomposition replaces continuous execution of the composite instruction with separately prompted subtasks and intermediate stopping decisions. This change in conditioning and execution boundaries may account for the initial decrease before adaptation. After three evolution rounds, both systems outperform their reproduced base policies on all three metrics, demonstrating gains beyond recovering the initial execution-only performance.

The gains also extend beyond the tasks used for evolution. Relative to the execution-only configurations, Composite-Unseen success rises from 4.7\% to 6.1\% with GR00T N1.5 and from 4.2\% to 7.1\% with RLDX-1. These tasks are excluded from both RoboCasa365 pretraining and \sees evolution. Both backbones improve throughout evolution and ultimately outperform their base policies on Composite-Unseen, demonstrating transfer to unseen tasks.
\vspace{-10pt}

\subsection{Self-Evolution on LIBERO}
\label{sec:libero-object}

LIBERO tasks are single-stage and require neither decomposition nor learned termination for outcome attribution. We therefore use LIBERO as a controlled setting that isolates the adaptation half of \sees: failure-guided target selection, joint training of a shared family adapter, and validation-based checkpoint acceptance. Each LIBERO task is treated as one atomic task, and the ten tasks of a suite form one family that shares a single adapter. This setting also tests whether the adaptation mechanism transfers beyond RoboCasa365 to a different simulator, task distribution, and set of publicly released starting checkpoints~\citep{yu2025rlinf}.

Each suite undergoes three evolution rounds under a deliberately limited budget: 80--96 outer PPO steps in total, updating only a rank-8 LoRA in the action head. In each round, only up to three failure-prone tasks are added as targets, so six to seven of the ten tasks receive RL training by the end of evolution. Evaluation pairs 15 held-out initial states with five inference seeds per task, yielding 75 trials per task and 750 per policy in each suite. Initial and evolved policies use identical state--seed pairs to control for differences in starting conditions and policy sampling. These states are excluded from target selection, training, and checkpoint validation (Appendix~\ref{app:libero-object}).

Evolution improves overall success in all four suites, raising the cross-suite average from 53.10\% to 65.40\% (Tab.~\ref{tab:libero-evolution}), and on LIBERO-Object every one of the ten tasks improves (Tab.~\ref{tab:libero-object-per-task}). The gains are not confined to the selected targets. Sibling tasks never participate in PPO, yet their success increases in every suite. In Object, Goal, and Long, siblings already start at 82--88\% success, leaving limited headroom, so these gains indicate positive transfer through the shared family adapter rather than improvement restricted to the trained bottlenecks. Rather than maximizing absolute success, which full-suite RL with larger budgets can achieve~\citep{yu2025rlinf}, these results show that concentrating a small RL budget on failure-selected tasks improves the entire suite.
\begin{table}[t]
  \caption{
    \textbf{LIBERO results after three evolution rounds.}
    Success rates (\%) compare the initial SFT policy with
    the policy retained after evolution.
    Bottleneck tasks are selected for PPO in at least one evolution
    round. Sibling tasks share their family adapter but never participate in PPO.
    Overall success covers all ten tasks in each suite.
    Arrows indicate initial-to-evolved performance.
  }
  \label{tab:libero-evolution}
  \centering
  \small
  \setlength{\tabcolsep}{6pt}
  \renewcommand{\arraystretch}{1.05}
  \begin{tabular*}{\linewidth}{@{\extracolsep{\fill}}llccc@{}}
    \toprule
    \multirow[c]{2}{*}{Suite}
      & \multirow[c]{2}{*}{Task group}
      & \multicolumn{2}{c}{Group success}
      & \multirow[c]{2}{*}{Overall success} \\
    \cmidrule(lr){3-4}
      & & Initial SFT & After evolution & \\
    \midrule
    \multirow[c]{2}{*}{Object}
      & Bottleneck tasks
      & 59.24 & \textbf{82.86}
      & \multirow[c]{2}{*}{66.13\,\(\rightarrow\)\,\textbf{86.00}} \\
      & Sibling tasks
      & 82.22 & \textbf{93.33} & \\
    \midrule
    \multirow[c]{2}{*}{Spatial}
      & Bottleneck tasks
      & 40.44 & \textbf{49.78}
      & \multirow[c]{2}{*}{41.33\,\(\rightarrow\)\,\textbf{50.53}} \\
      & Sibling tasks
      & 49.33 & \textbf{57.33} & \\
    \midrule
    \multirow[c]{2}{*}{Goal}
      & Bottleneck tasks
      & 39.11 & \textbf{56.15}
      & \multirow[c]{2}{*}{43.73\,\(\rightarrow\)\,\textbf{59.47}} \\
      & Sibling tasks
      & 85.33 & \textbf{89.33} & \\
    \midrule
    \multirow[c]{2}{*}{Long}
      & Bottleneck tasks
      & 58.22 & \textbf{62.96}
      & \multirow[c]{2}{*}{61.20\,\(\rightarrow\)\,\textbf{65.60}} \\
      & Sibling tasks
      & 88.00 & \textbf{89.33} & \\
    \bottomrule
  \end{tabular*}
  \vspace{-15pt}
\end{table}

\vspace{-8pt}
\subsection{Family-Level Adaptation and Transfer}
\label{sec:family-transfer}
\vspace{-6pt}
We compare separate atomic-specific LoRAs with a shared
family-level LoRA on five related pick-and-place tasks.
Three bottleneck tasks are selected for adaptation. Each atomic-specific
LoRA is trained for 40 steps using initializations from its own
bottleneck task. The shared family LoRA is trained for 40 steps in total
using mixed initializations from all three bottleneck tasks. The remaining two tasks
are excluded from RL and evaluated as siblings.
This comparison matches the 40-step budget per adapter, while the shared
adapter jointly learns from all three bottleneck tasks within a single run.
For the atomic-specific baseline, bottleneck-task success averages
the three adapters' performance on their respective training tasks, while sibling-task success
averages across all adapter--sibling pairs.
Tab.~\ref{tab:family-rl-transfer} reports mean success
separately for these two groups, allowing us to distinguish
improvement on the training targets from transfer to
non-target tasks.

\begin{table}[b]
  \vspace{-16pt}
  \centering
  \caption{\textbf{Atomic-specific versus family-level RL adaptation.}
  Success rates (\%) are reported for the three selected bottleneck tasks
  and two sibling tasks excluded from RL.
  Overall success averages all five tasks with equal weight.}
  \label{tab:family-rl-transfer}
  \small
  \setlength{\tabcolsep}{5pt}
  \renewcommand{\arraystretch}{1.10}

  \begin{tabular*}{\linewidth}{@{\extracolsep{\fill}}lccc@{}}
    \toprule
    Method
      & \shortstack{Bottleneck tasks}
      & \shortstack{Sibling tasks}
      & \shortstack{Overall} \\
    \midrule
    Before RL
      & 62.7
      & 75.0
      & 67.6 \\
    \midrule
    Atomic-specific LoRA
      & 70.7
      & 72.0
      & 71.2 \\
    Family-level LoRA
      & \textbf{74.0}
      & \textbf{80.0}
      & \textbf{76.4} \\
    \bottomrule
  \end{tabular*}
\end{table}

In this controlled comparison, family-level adaptation achieves higher
success on both bottleneck and sibling tasks. Atomic-specific adaptation
improves its training targets but shows no positive mean transfer to sibling
tasks, supporting shared family adapters for cross-task reuse.

Fig.~\ref{fig:evolution-transfer} illustrates the same
failure-to-update cycle in a composite task.
The family update improves the targeted bottleneck,
sibling-task performance, and full-composite success.
Target-task success is measured from reconstructed RL initializations, while sibling-task and composite-task evaluations assess transfer to related skills and improvements in end-to-end execution, respectively.

\begin{table*}[htbp]
  \centering
  \caption{\textbf{Ablation of adaptation-target selection strategies on
  \texttt{ScrubCuttingBoard}.}
  Results report full-task success on 50 held-out scenes after two
  evolution rounds. All strategies use the same base policy and
  40-step RL budget in each round.}
  \label{tab:selection-ablation}
  \small
  \setlength{\tabcolsep}{8pt}
  \renewcommand{\arraystretch}{1.10}
  \begin{tabular*}{\textwidth}{
    @{\extracolsep{\fill}}lccc@{}
  }
    \toprule
    Selection strategy
      & Round 1 target
      & Round 2 target
      & Held-out full-task success \\
    \midrule
    Base policy
      & -- & -- & 9/50 (18\%) \\
    Random
      & Release
      & Navigate
      & 14/50 (28\%) \\
    Highest-success stage
      & Pick up
      & Release
      & 17/50 (34\%) \\
    Bottleneck (ours)
      & Scrub
      & Navigate
      & \textbf{32/50 (64\%)} \\
    \bottomrule
  \end{tabular*}
  \vspace{-12pt}
\end{table*}

\subsection{Ablation of Bottleneck Selection}
\label{sec:bottleneck-ablation}
\vspace{-6pt}

We test whether prioritizing the most frequent blocker yields
more full-task improvement under a fixed adaptation budget.
On \texttt{ScrubCuttingBoard}, we compare our failure-based selection
with random selection and a highest-success control that allocates
training to the currently most reliable stage. All strategies start
from the same base policy and follow the RoboCasa365 evolution protocol,
using two evolution rounds with 40 PPO steps and the same
checkpoint-validation budget per round.

Tab.~\ref{tab:selection-ablation} shows that bottleneck selection raises
full-task success from 18\% to 64\%, compared with 28\% for random
selection and 34\% for the highest-success control. The selected
bottleneck shifts from \textit{Scrub} in the first round to
\textit{Navigate} in the second as execution outcomes are recomputed
after the update. On this controlled task, directing the same adaptation
budget toward the current bottleneck produces substantially larger
full-task gains than either alternative selection strategy.

\vspace{-5pt}
\section{Real-World Feasibility and Future Work}
\label{sec:limitations}
\label{sec:future-work}
\vspace{-8pt}
All signals required by the execution stage of \sees are available or can be inferred during physical-robot deployment. The planner and VLA policy operate on task instructions and robot observations, the termination head infers subtask completion from the same observations, and bottleneck selection aggregates the resulting execution outcomes.
In the evolution stage, the main additional requirement is an interactive simulator that recreates the execution contexts of selected bottlenecks.

Recent real-to-sim methods provide complementary modules for constructing such environments. URDFormer reconstructs articulated scenes from images~\citep{chen2024urdformer}, while ACDC automatically creates interactive scenes that preserve the geometric and semantic affordances of real environments without requiring exact replicas~\citep{dai2025digitalcousins}. SplatSim, RoboGSim, and RE$^3$SIM combine reconstructed visual representations with physics simulation to support policy learning and transfer~\citep{qureshi2024splatsim,li2025robogsim,han2026re3sim}. Because \sees targets individual bottlenecks, it only needs to reconstruct the objects and interactions relevant to the selected subtask rather than the entire scene. Scene scans could supply task-relevant geometry, and execution logs could estimate the configuration before the subtask begins. These estimates would initialize simulated practice, randomized over pose and dynamics to absorb reconstruction error. Within such environments, the task builder could generate success functions from atomic instructions and the simulator's state-query interface, as in Sec.~\ref{sec:rl-task-construction}, and use them to define sparse RL rewards. RialTo and CASHER show that learning in reconstructed environments can improve real-world policies~\citep{torne2024rialto,torne2025casher}, suggesting that family adapters could likewise be updated off-robot. Since \sees updates only family adapters while keeping the pretrained backbone frozen, candidate adapters can be screened on hardware before being reused across related tasks. Future work will integrate these components and evaluate the complete loop, testing whether targeted adaptation yields sustained improvements in both bottleneck skills and long-horizon execution.

\Needspace{12\baselineskip}
\section{Conclusion}
\vspace{-8pt}
We presented \sees, a failure-guided self-evolution system that turns recurring atomic bottlenecks in long-horizon VLA execution into targeted simulation RL updates. Using restored pre-task states and shared family adapters, \sees improves related atomic capabilities without additional expert demonstrations. On RoboCasa365, the evolved policies outperform their respective base policies across both backbones, including on Composite-Unseen tasks. Across all four LIBERO suites, evolution also benefits tasks excluded from RL adaptation. Controlled analyses further support termination-based failure attribution, bottleneck selection, and family-level adaptation, demonstrating the effectiveness of the closed evolution loop.
\clearpage

\bibliographystyle{iclr2027_conference}
\bibliography{references}

\appendix

\clearpage
\section{Execution and Routing Details}
\label{app:vlm-decomposition}

\begin{figure}[H]
  \centering
  \includegraphics[width=\linewidth]{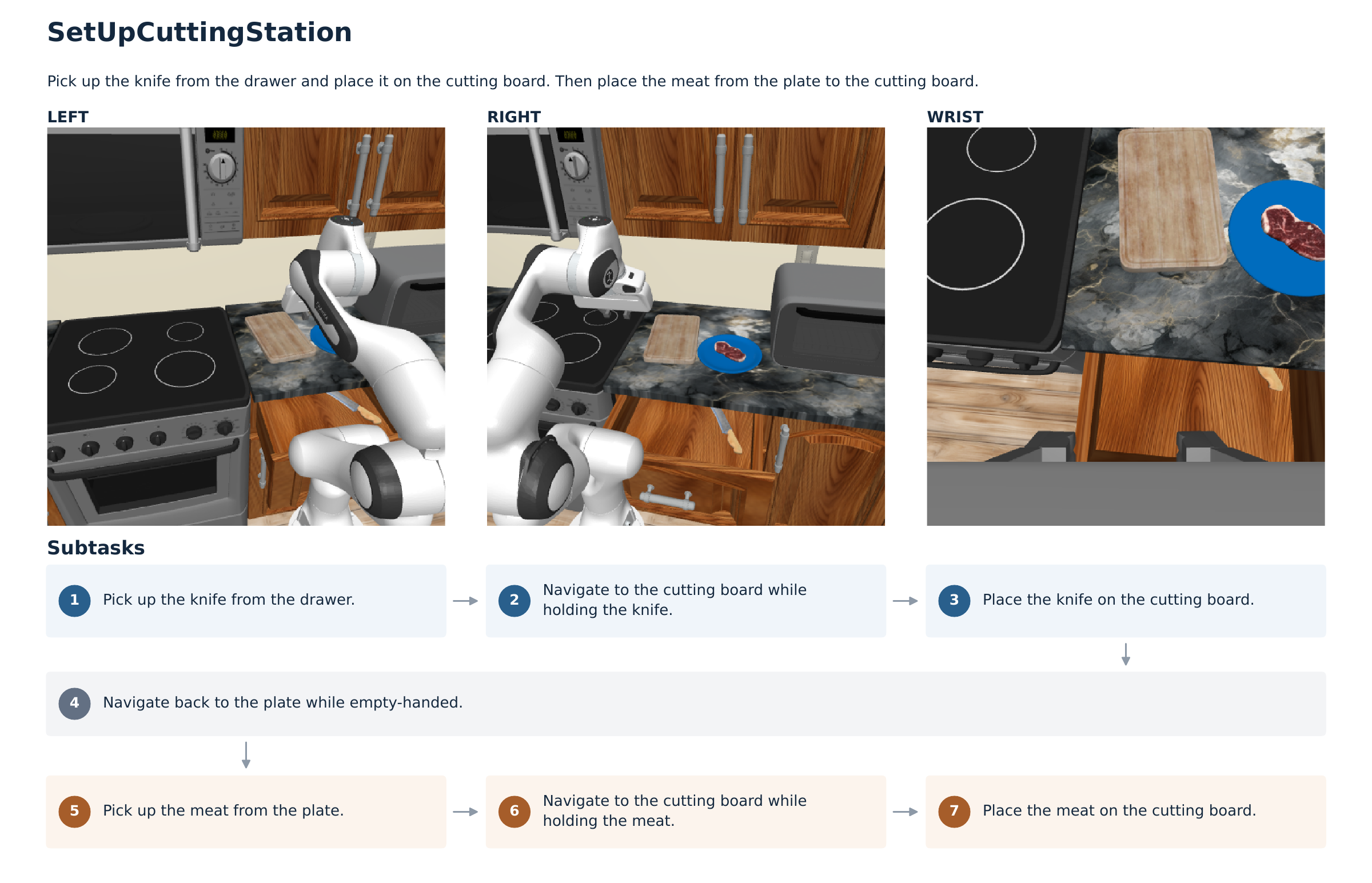}
  \caption{\raggedright\textbf{VLM task decomposition.} The generated seven-step plan for \texttt{SetUpCuttingStation}.}
  \label{fig:vlm-decomposition-case}
\end{figure}

\noindent\textbf{Setup.}
We instantiate the planner with a frozen Gemma 4 31B IT
model~\citep{gemmateam2026gemma4}.
Given the task name, complete instruction, and synchronized
left, right, and wrist images of the initial scene,
the model produces an ordered sequence of atomic
instructions in JSON format.
The complete prompt used by the VLM planner is provided in the
supplementary file \mbox{\texttt{task\_decomposition.md}}.
A single prompt defines stage boundaries,
operation grouping, object relations, and ordering
conventions through explicit rules.
We use greedy decoding without parameter fine-tuning,
multiple-candidate selection, or post-generation
semantic repair.
Fig.~\ref{fig:vlm-decomposition-case} illustrates the resulting decomposition
for \texttt{SetUpCuttingStation} from initial observations and the task instruction.

\noindent\textbf{Evaluation.}
We evaluate the planner on 32 RoboCasa365 composite task
types with two initial states each, totaling 64 cases.
A plan is considered correct only when its stage count,
ordering, actions, object relations, and navigation
boundaries agree with the reference specifications.
Semantically equivalent wording is accepted.
Agreement is assessed through AI-assisted review,
with any critical mismatch or invalid output counted
as a failure.
The planner achieves 63/64 whole-plan agreement
(98.44\%), and all outputs are valid JSON.
The sole error identifies the correct tool but assigns
its source to the counter rather than an open drawer.

\noindent\textbf{Atomic specifications and skill-family routing.}
For reproducible RoboCasa365 evaluation, we freeze planner outputs as benchmark-normalized atomic specifications. Each specification is assigned to one of 11 semantic families: opening articulated fixtures, closing articulated fixtures, operating appliances, controlling sink water, grasping or retrieving objects, placing objects in portable receptacles, placing objects on surfaces or other objects, inserting objects into fixed fixtures or racks, arranging multiple objects relationally, pouring or rinsing, and goal-conditioned navigation. These predefined families determine parameter sharing rather than replacing the benchmark's atomic task definitions.

A language agent assigns each specification to a family from its instruction and the skill-library descriptions. Assignments are cached with the library version. Routing selects the family's latest registered adapter rather than the nearest previously trained atomic-task adapter. The base VLA is used when that family has no adapter. The selected policy remains active for the full atomic attempt.

\noindent\textbf{Outcome statistics.}
For composite task $c$ and atomic stage $i$, the system records reach count $R_{c,i}$, success count $S_{c,i}$, and attributed failure count $F_{c,i}$, with $R_{c,i}=S_{c,i}+F_{c,i}$. Counts use the learned termination rule described in Sec.~\ref{sec:termination}. With the same rollout budget for every composite task, selecting the largest failure count is equivalent to selecting the largest share of task rollouts attributed to a stage. Taking one bottleneck per composite task avoids allocating all targets to a small set of frequently occurring tasks.

\section{Termination-Head Training and Failure-Attribution Audit}
\label{app:termination-audit}
\noindent\textbf{Head-only termination training.}
Starting from a pretrained VLA checkpoint, we attach a prediction head to its action-token features. The head has termination and progress outputs and consists of layer normalization, a hidden layer of width 512 with GELU activation, and a two-output projection. All existing VLA parameters, including the action head, are frozen throughout this stage. Only the added prediction head is optimized, with the head-only objective
\begin{equation}
\mathcal{L}_{\mathrm{head}}
=
0.10\,\mathcal{L}_{\mathrm{term}}^{\mathrm{SFT}}
+0.05\,\mathcal{L}_{\mathrm{progress}}.
\label{eq:termination-warmup}
\end{equation}
Termination supervision uses masked binary cross-entropy with soft targets that increase linearly from zero to one over a 32-frame window centered on the annotated completion frame. Progress supervision uses masked L1 regression against normalized progress toward that frame. The termination loss uses a dynamically balanced positive-class weight clipped to $[1,50]$. These losses optimize only the added head, not the existing VLA action policy. This head-only preparation is distinct from the subsequent RL stage.

\noindent\textbf{Completion labels and inference rule.}
For an atomic demonstration of length $L$, we use its final recorded
action frame, $c=L-1$, as the completion frame. For segmented composite
demonstrations, we use the recorded subtask-completion boundary.
These completion frames anchor the soft termination targets described
above. During composite execution, the head predicts a termination
probability for each action in the generated chunk. Completion is
declared when this probability is at least $0.95$ for three consecutive
environment steps. The counter is updated after each executed action,
carried across chunk boundaries, and reset whenever the probability
falls below the threshold or a new subtask begins. Once completion is
declared, the remaining actions in the current chunk are discarded
and execution advances to the next subtask.

\noindent\textbf{Audit protocol.}
We compare two decision rules, Terminate and VLM, on 100 composite
rollouts (50 per task), comprising 243 atomic attempts. The VLM is the open-weight Qwen3-VL-32B-Instruct model, deployed locally
in BF16 on one 80GB A800 GPU. At the end of every action chunk, it receives
the current atomic instruction and synchronized left, right, and wrist RGB
observations, then returns a structured Boolean completion decision. The simulator oracle
controls composite execution, while auxiliary rollouts continue each
completed subtask to evaluate the rules on shared trajectories within the
original horizon. We report pooled outcome Macro-F1, false success
($FP/N$), mean absolute termination-time error, and rollout-level
first-failure attribution accuracy. Timing error compares the first
success-declaration times of each rule and the oracle, assigning the
horizon endpoint when no success is declared. Matching \texttt{NONE}
attribution labels count as correct.

\begin{table}[H]
  \centering
  \small
  \setlength{\tabcolsep}{6pt}
  \caption{\textbf{Atomic termination and failure-attribution evaluation.}
  Both rules are evaluated on shared trajectories with
  oracle-controlled composite execution.
  Values are percentages except mean absolute termination-time
  error (control steps).}
  \label{tab:outcome-audit}

  \begin{tabular}{lcccc}
    \toprule
    Decision rule
      & \shortstack{Outcome\\Macro-F1 $\uparrow$}
      & \shortstack{False\\success $\downarrow$}
      & \shortstack{Termination-time\\error (steps) $\downarrow$}
      & \shortstack{Failure attrib.\\accuracy $\uparrow$} \\
    \midrule
    Simulator oracle
      & 100.0 & 0.0 & 0.0 & 100.0 \\
    \midrule
    Terminate
      & \textbf{94.0} & \textbf{3.3} & \textbf{10.8} & \textbf{92.0} \\
    VLM
      & 59.9 & 17.7 & 112.6 & 45.0 \\
    \bottomrule
  \end{tabular}
\end{table}

Terminate outperforms VLM on all four metrics
(Tab.~\ref{tab:outcome-audit}).
It achieves higher outcome Macro-F1 (94.0\% vs.\ 59.9\%)
and first-failure attribution accuracy (92.0\% vs.\ 45.0\%),
with fewer false successes (3.3\% vs.\ 17.7\%) and lower
mean absolute termination-time error (10.8 vs.\ 112.6 steps).
These results favor policy-native termination in this audit.
They do not establish improvements in online composite-task
success or general limitations of VLM-based monitoring.

\section{RL Task Construction and Family-Level Adaptation Details}
\label{app:rl-construction}

This appendix expands the task-construction and family-update procedures in Secs.~\ref{sec:rl-task-construction} and~\ref{sec:family-evolution}. It follows the RoboCasa365 execution-to-adaptation pipeline: restoring encountered states, generating success functions and rewards, sampling family-level training tasks, optimizing the policy, and registering validated checkpoints.

\subsection{State Collection and Restoration}
\label{app:state-restoration}

For each selected atomic task, the builder collects initializations recorded immediately before that task in composite rollouts. Both successful and failed attempts contribute pre-task states. Each record contains the atomic instruction, simulator state, scene description, and episode metadata, including task-dependent state such as appliance status and timers.

Restoration reinstates the episode configuration and scene, loads the saved simulator state, updates the simulation, and restores task-dependent state before obtaining observations. We use the captured states without additional temporal jitter. Initialization diversity comes from the different contexts encountered during composite execution.

An initialization is eligible for RL when its required records are complete, restoration succeeds, and the generated success function from Appendix~\ref{app:criterion-generation} returns false. These unfinished instances form $\mathcal{I}_i$ in Eq.~\eqref{eq:rl_task}. The builder pairs each valid initialization with its instruction, success function, and collection source, then assembles the corresponding RL configuration.

Training resets may retry another state within the assigned task's pool under a fixed retry budget. Exhausting the budget produces an initialization error rather than a rollout from an invalid state. During evaluation, retries preserve the prescribed initial-state assignment. An unresolved reset leaves the scheduled trial incomplete, distinct from policy failure, and prevents that comparison from being used for checkpoint registration.

\subsection{Success-Checking Functions and Sparse Rewards}
\label{app:criterion-generation}

For each selected atomic task $i$, a script fills a predefined prompt template with its instruction $l_i$, the identifiers of relevant objects and fixtures, and documentation for simulator state-query functions. We use GPT-5.6-Sol~\citep{openai2026gpt56sol} as the code-generating LLM. Given the
populated prompt, it produces a Python function $C_i(s)$ that returns
true when the task is complete and false otherwise. The function can combine simulator queries to check the required object states and relations. For example, a grasping function checks whether the robot is grasping the target object through the simulator's object-grasping interface.

The generated function directly supplies both initialization checks and sparse RL rewards. During PPO rollouts, $C_i$ must hold for three consecutive environment steps before success is confirmed. The latched indicator and reward follow Eq.~\eqref{eq:rl_reward}, giving a reward of one at the first confirmed completion and zero otherwise. We use no additional dense reward shaping or learned reward model.

During RL, the generated success function determines task success and episode completion. The termination head receives auxiliary supervision but does not control RL rollout stopping or supply the reward. During composite execution, the learned termination head instead determines atomic-task handoff, as described in Sec.~\ref{sec:termination}.

\noindent\textbf{Agreement with official success predicates.}
We further evaluate the executable success predicates generated by the LLM on 900 Atomic-Seen rollouts (18 tasks, 50 seeds per task) from the official GR00T N1.5 checkpoint. The generated and official predicates are evaluated on every state of the same trajectory, with success confirmed after three consecutive positive states. Frame accuracy measures per-state agreement, while balanced accuracy equally weights agreement on successful and unsuccessful states to account for their imbalance. Episode-level agreement measures whether the two predicates reach the same final success outcome for an entire rollout. As shown in Tab.~\ref{tab:atomic-seen-predicate-agreement}, the generated predicates achieve 97.8\% frame accuracy, 98.2\% balanced accuracy, and 95.9\% episode-level agreement with the official environment predicates; 13 of 18 tasks have exact episode-level agreement.
These results support the accuracy of success predicates generated from task instructions and documented simulator APIs on the evaluated Atomic-Seen tasks.

\begin{table*}[t]
  \centering
  \caption{\textbf{Agreement between LLM-generated and official RoboCasa365 success predicates on Atomic-Seen.}
  Values are task-macro percentages, evaluated on identical GR00T N1.5 checkpoint-120000 trajectories.}
  \label{tab:atomic-seen-predicate-agreement}
  \small
  \begin{tabular*}{\textwidth}{@{\extracolsep{\fill}}lcccc@{}}
    \toprule
    & Frame accuracy & Balanced accuracy & Recall & Episode agreement \\
    \midrule
    LLM-generated predicate
    & 97.8
    & 98.2
    & 96.4
    & 95.9 \\
    \bottomrule
  \end{tabular*}
\end{table*}

\subsection{Family-Level Task Sampling}
\label{app:family-sampling}
Current bottlenecks and previously selected tasks are grouped by their skill-family assignments. From the second evolution round onward, each active family trains on their union, with duplicate tasks included only once. Training samples a task and then one of its restored states to collect fresh rollouts with the current family policy. Historical tasks are revisited through new interaction rather than stored trajectories.

Let $N_{i,k}$ be the number of unique valid initial states for task $i$ in round $k$, and $k_i^{\mathrm{first}}$ its first-selection round. Its sampling weight and nominal within-family sampling probability are
\begin{equation}
w_{i,k}=\sqrt{N_{i,k}}\,2^{-(k-k_i^{\mathrm{first}})},
\qquad
p_{i,k}=
\frac{w_{i,k}}{\sum_{j\in\mathcal{T}_{f,k}}w_{j,k}},
\label{eq:tempered-replay}
\end{equation}
where $\mathcal{T}_{f,k}$ contains the family's valid training tasks. This balances initialization coverage with gradually reduced emphasis on older tasks. Neither reselection nor replay resets task age. Weights remain fixed within each round and affect environment initialization, not PPO loss weighting. Starts are sampled uniformly within each task.

\subsection{PPO Optimization}
\label{app:policy-optimization}
We use the PPO training code from RLinf~\citep{yu2025rlinf} for our RL updates.
For the low-rank parameterization in Eq.~\eqref{eq:family-lora}, $W_0\in\mathbb{R}^{d_{\mathrm{out}}\times d_{\mathrm{in}}}$, $A_f\in\mathbb{R}^{r\times d_{\mathrm{in}}}$, and $B_f\in\mathbb{R}^{d_{\mathrm{out}}\times r}$.
For the termination-aware configuration, LoRA adapts the action head with rank 64, scaling factor 128, and zero dropout. The termination and value heads remain trainable and are saved with the adapter, while the pretrained vision and language backbone remains frozen. The actor uses stochastic flow transitions during PPO, retaining sampled denoising states and their rollout log-probabilities. Updates recompute likelihoods for the same transitions to optimize Eq.~\eqref{eq:family-ppo}. These likelihoods describe the sampled stochastic transitions rather than an exact marginal likelihood of the final action.

The training objective combines the clipped PPO objective, value regression, and termination supervision:
\begin{equation}
\mathcal{L}_{\mathrm{RL}}
=
-J_f^{\mathrm{PPO}}
+\mathcal{L}_{V}
+0.05\,\mathcal{L}_{\mathrm{term}}^{\mathrm{RL}}.
\label{eq:termination-aware-rl}
\end{equation}
Termination targets are derived from the RL environment's atomic-success signal. Within each action chunk, the target becomes positive at the first successful transition and remains positive thereafter. Padding and post-terminal positions are masked while retaining the first terminal transition. Time-limit truncation alone does not produce a positive completion target. We use masked binary cross-entropy with logits, with the negative-to-positive target ratio determining a positive-class weight clipped to $[1,50]$. No separate progress-prediction loss is added during RL. This auxiliary training during evolution is distinct from the head-only preparation described in Appendix~\ref{app:termination-audit}.

The termination-aware PPO configuration uses 32 parallel environments on two GPUs, 40 outer training steps per active family per evolution round, and checkpoint saving every ten steps. Each outer step uses four rollout epochs and two update epochs, with a global batch size of 240 and a micro-batch size of 12. Actor and value-head learning rates are $6\times10^{-6}$ and $10^{-4}$, respectively. PPO uses a clipping range of $0.2$, discount factor $0.99$, and GAE parameter $0.95$. Value regression uses a clipped Huber loss with clipping range $0.2$ and threshold $10$. Gradient clipping is set to $1.0$, while entropy regularization and the KL penalty are disabled. The action chunk length is 16, the number of denoising steps is four, and the episode horizon is 720 environment steps. The policy seed is 42 and the environment seed is 0.

\subsection{Checkpoint Validation and Reuse}
\label{app:checkpoint-validation}
After each family update, we evaluate the policy before the update and the saved checkpoints from steps 10, 20, 30, and 40. For each selected bottleneck atomic task, we randomly sample 5 valid initial states. The valid states are used to evaluate every checkpoint and the policy before the update under matched settings. Each policy is evaluated five times from these states.

For each checkpoint $j$, we compute success on every selected bottleneck task and average equally across these tasks. Its gain over the pre-update policy is $\Delta_T^{(j)}$, as defined in Sec.~\ref{sec:family-evolution}. We register the checkpoint with the largest positive gain according to Eq.~\eqref{eq:target-checkpoint-selection}. If no checkpoint improves this mean success rate, we retain the current family policy. The registered checkpoint is shared by all tasks routed to that family and initializes its subsequent adaptation.

The registry stores the family identity, LoRA version, contributing atomic tasks, initialization source, reward definition, and validation results. Subsequent composite executions use the updated family policies and supply failure statistics for the next evolution round. Atomic handoff follows the learned termination rule, while the benchmark predicate determines final composite success. This full-task evaluation measures system-level effects rather than imposing another registration threshold. Composite-Unseen remains excluded from adaptation, checkpoint selection, and registration.

\section{LIBERO Experimental Details}
\label{app:libero-object}

\noindent\textbf{Protocol.}
We evaluate policy evolution independently on the Object, Spatial, Goal,
and Long suites of LIBERO. Each suite starts from the corresponding GR00T N1.5
SFT checkpoint released by RLinf~\citep{yu2025rlinf} and adapts LoRA modules in the action head using PPO.
For this evaluation, we treat the ten tasks within each suite as one
skill family and use a single family adapter shared across them.
Each task is executed using its suite's current family policy.
Policies and adapters are not transferred between suites.

For every suite, each task has 50 predefined initial states partitioned
into 20 training states, five approval states for each of three rounds,
and 15 sealed final-evaluation states. These partitions are mutually
disjoint. The sealed states are excluded from failure-based target
selection, PPO training, candidate approval, and checkpoint selection.
Final evaluation pairs each sealed state with five inference seeds,
giving 75 trials per task and 750 trials per policy in each suite.
The initial and evolved policies use identical state--seed pairs.
The five seeds characterize inference variability for a single
evolution run per suite, rather than five independent training runs.
\begin{table}[H]
  \caption{\textbf{Per-task sealed LIBERO-Object results.} Each entry contains 75 trials using 15 sealed states and five inference seeds. Tasks marked ``not selected'' never enter PPO training in the accepted chain.}
  \label{tab:libero-object-per-task}
  \centering
  \scriptsize
  \setlength{\tabcolsep}{5pt}
  \begin{tabular}{lcccc}
    \toprule
    Task & Initial SFT & After evolution & Gain (pp) & First selected \\
    \midrule
    Alphabet soup & 35/75 (46.67\%) & 56/75 (74.67\%) & +28.00 & R2 \\
    Cream cheese & 64/75 (85.33\%) & 72/75 (96.00\%) & +10.67 & R2 \\
    Salad dressing & 27/75 (36.00\%) & 56/75 (74.67\%) & +38.67 & R3 \\
    Barbecue sauce & 50/75 (66.67\%) & 64/75 (85.33\%) & +18.67 & R3 \\
    Ketchup & 50/75 (66.67\%) & 65/75 (86.67\%) & +20.00 & Not selected \\
    Tomato sauce & 31/75 (41.33\%) & 57/75 (76.00\%) & +34.67 & R1 \\
    Butter & 68/75 (90.67\%) & 72/75 (96.00\%) & +5.33 & Not selected \\
    Milk & 52/75 (69.33\%) & 62/75 (82.67\%) & +13.33 & R3 \\
    Chocolate pudding & 67/75 (89.33\%) & 73/75 (97.33\%) & +8.00 & Not selected \\
    Orange juice & 52/75 (69.33\%) & 68/75 (90.67\%) & +21.33 & R2 \\
    \midrule
    \textbf{All tasks} & \textbf{496/750 (66.13\%)} & \textbf{645/750 (86.00\%)} & \textbf{+19.87} & -- \\
    \bottomrule
  \end{tabular}
\end{table}

\noindent\textbf{Evolution rounds.}
Each suite undergoes three evolution rounds. At each round, we prioritize
failure-prone tasks using rollouts from the current policy on the
training split and select up to three new target tasks. Previously
selected tasks are excluded from new target selection until all tasks
have been covered.

From the second round onward, the training pool is the union of the
newly selected tasks and all previously selected tasks, including those
selected in rounds whose candidates were rejected. If task $i$ was most
recently selected in round $r_i$, its unnormalized sampling weight in
round $r$ is
\[
    w_i^{(r)} = 0.5^{\,r-r_i}.
\]
We normalize these weights over the training pool, sample a task,
and then uniformly sample one of its 20 training initial states.
Weights remain fixed within a round, and replaying a historical task
does not update $r_i$. This mechanism changes environment initialization
probabilities, not PPO loss weights. All trajectories are collected
anew using the current policy. Trajectories from earlier rounds are
not reused.

Each candidate is compared with the policy at the start of the
current round on disjoint approval states, using the same five
inference seeds.
A candidate is accepted only if it strictly improves average success on the newly selected tasks.
Performance on historical tasks
and the full suite is recorded but does not impose an additional
acceptance threshold. Subsequent rounds initialize from the latest
accepted policy, and final evaluation uses the policy retained after
the third round.

\noindent\textbf{Optimization.}
The action-head LoRA uses rank 8, alpha 8, and zero dropout.
Training uses 16 parallel environments on two GPUs, a global batch
size of 512, a micro-batch size of 8, four rollout epochs, one update
epoch, an actor learning rate of $5\times10^{-6}$, a critic learning
rate of $10^{-5}$, and gradient clipping at 0.5.
The optimizer is reinitialized in each round, with 48 critic-only
optimizer updates before actor updates begin.

Rounds use 32 outer training steps, except for the initial single-task
round, which uses 16. The resulting three-round candidate-training
budgets are 80, 96, 96, and 96 outer steps for Object, Spatial, Goal,
and Long, respectively. These budgets include rejected candidates
and therefore need not equal the training incorporated into the
final accepted policy. An outer step includes rollout collection
and optimization and is not a single gradient update.

\noindent\textbf{Inference settings.}
The episode horizons are 240, 240, 320, and 480 environment steps for
Object, Spatial, Goal, and Long, respectively. The corresponding
action-chunk lengths are 5, 5, 5, and 10, with four denoising steps
for all suites. Final evaluation uses ordered initial states,
30 parallel environments on two GPUs, and five evaluation rollout
epochs per inference seed, with automatic reset disabled.

\noindent\textbf{Example results.}
Tab.~\ref{tab:libero-object-per-task} presents LIBERO-Object as an
example of the per-task evaluation breakdown. It reports success
counts and rates before and after evolution, gains in percentage
points, and the round in which each task was first selected for
adaptation.

\section{Controlled Qualitative Replays}
\label{sec:qualitative-replays}

\begin{figure}[tbp]
  \centering
  \includegraphics[width=\linewidth]{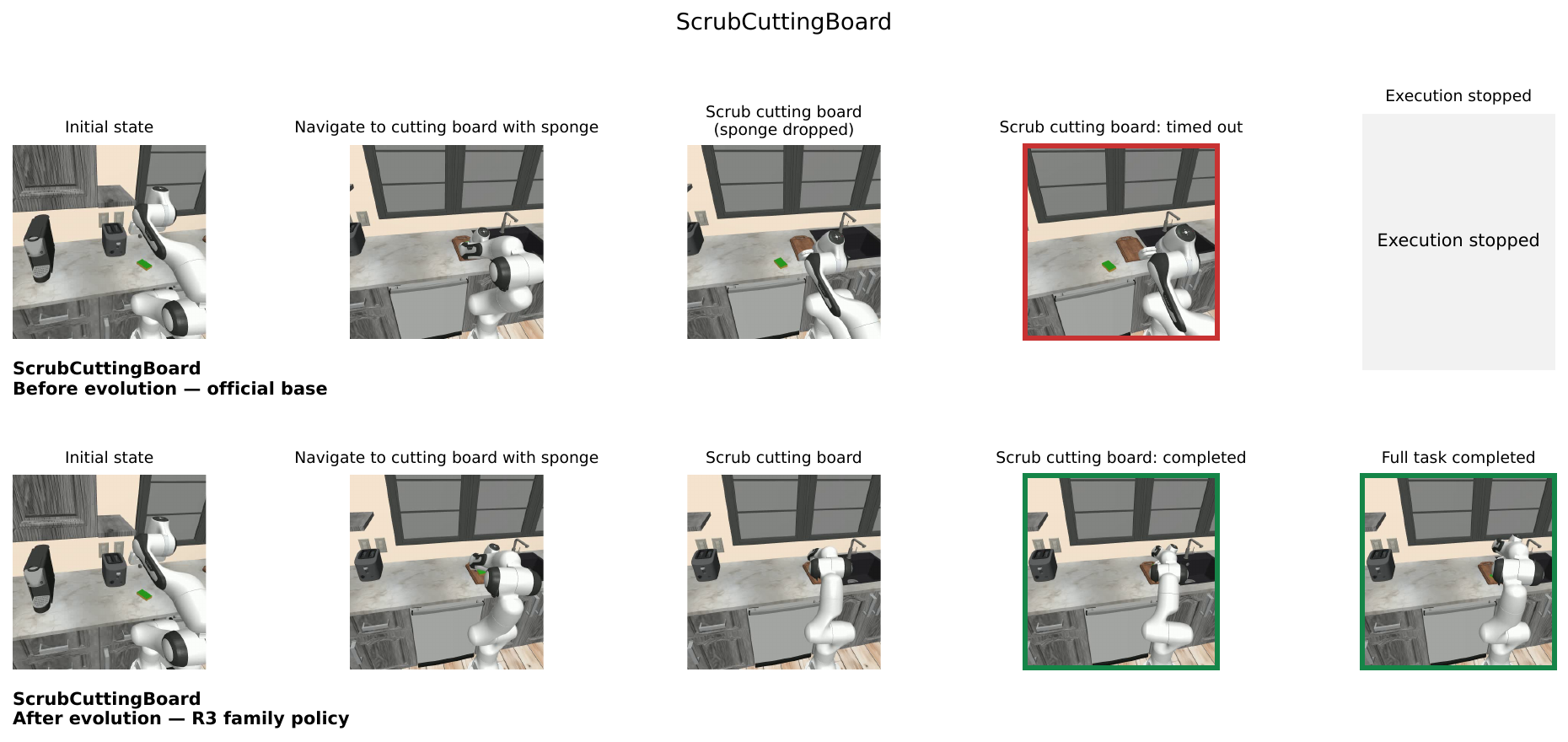}
  \caption{
    \textbf{Bottleneck recovery and continued execution.}
    In this paired \emph{ScrubCuttingBoard} replay, the GR00T baseline
    loses the sponge and times
    out at the scrubbing stage (top).
    The full validated round-3 candidate routing completes
    scrubbing and the remaining stages (bottom).
    The evolved row uses the complete family routing,
    not a single-adapter replacement.
    The final upper panel marks stopped execution and is not
    a recorded downstream frame.
  }
  \label{fig:qualitative-scrub}
\end{figure}

\begin{figure}[p]
  \centering
  \includegraphics[
    width=\linewidth,
    height=0.72\textheight,
    keepaspectratio
  ]{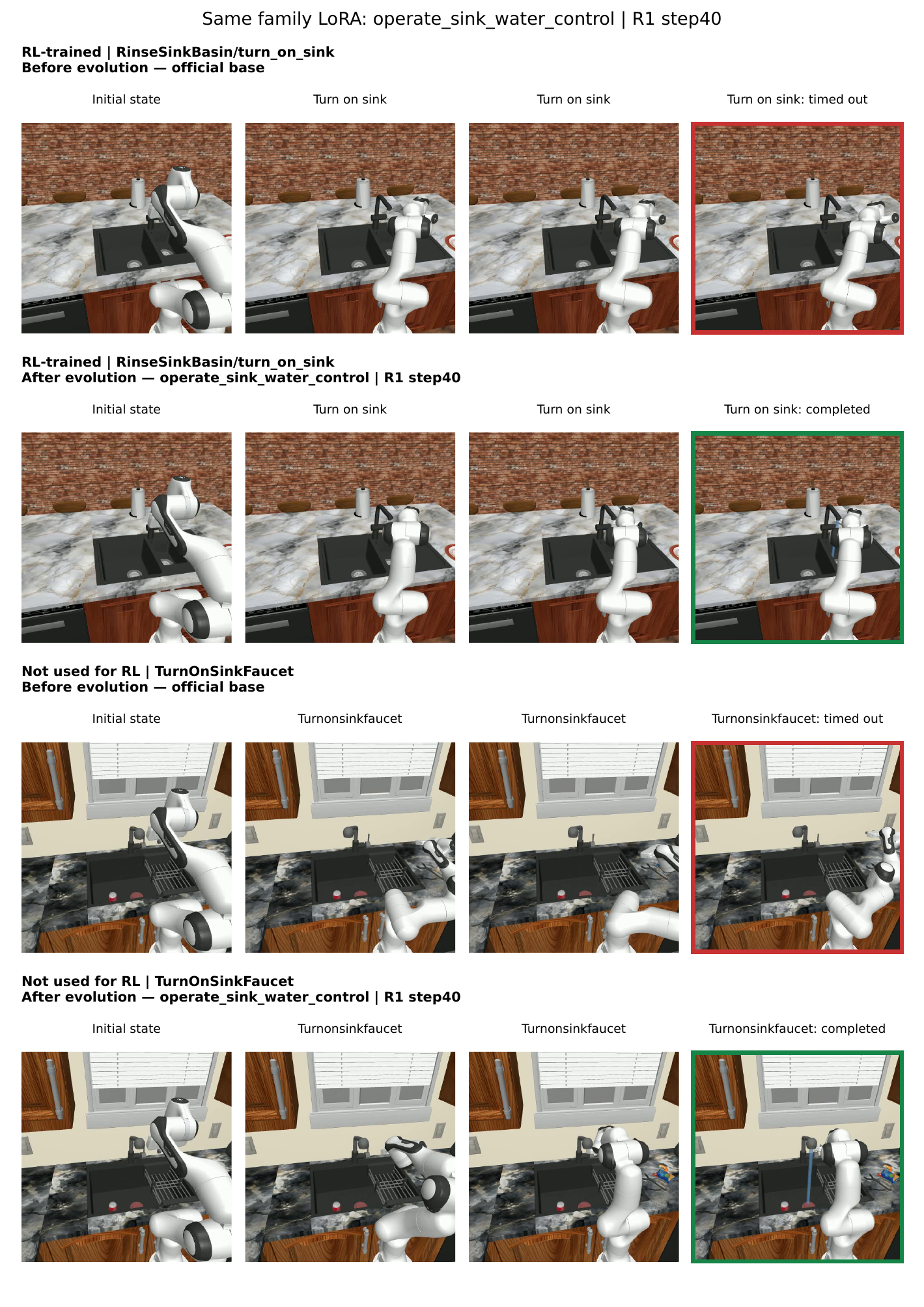}
  \caption{
    \textbf{Transfer with one shared sink-control adapter.}
    The baseline is compared with the same
    round-1 sink-control LoRA in both cases:
    the RL-trained
    \emph{RinseSinkBasin/turn\_on\_sink} subtask (upper pair)
    and the native \emph{TurnOnSinkFaucet} atomic task
    (lower pair).
    The base fails and the shared adapter succeeds in each
    displayed pair.
  }
  \label{fig:qualitative-sink-transfer}
\end{figure}

Figs.~\ref{fig:qualitative-scrub}
and~\ref{fig:qualitative-sink-transfer}
show independently recorded, controlled replays of existing
checkpoints under a separate visualization protocol.
Each before/after pair restores the same full initial state
and uses a matched inference seed.
The figures show one representative pair from two
pre-specified repeats per case.

\end{document}